\documentclass{article}

\usepackage[english]{babel}

\usepackage[letterpaper,top=2cm,bottom=2cm,left=3cm,right=3cm,marginparwidth=1.75cm]{geometry}

\usepackage{amsmath}
\usepackage{graphicx}
\usepackage[colorlinks=true, allcolors=blue]{hyperref}
\usepackage{appendix}
\usepackage[utf8]{inputenc} 
\usepackage[T1]{fontenc}    
\usepackage{multirow}

\usepackage[most]{tcolorbox}
\usepackage{xcolor}
\usepackage{lipsum}
\definecolor{boxred}{RGB}{150,0,0}
\definecolor{boxpurple}{RGB}{60,0,150}

\title{Does This Look Familiar to You? 

Knowledge Analysis via Model Internal Representations}
\author{Sihyun Park \\  {psh0430@dongguk.edu} }

\begin{document}

\maketitle
\begin{abstract}
Recent advances in large language models (LLM) have been driven by pretraining, supervised fine tuning (SFT), and alignment tuning. Among these, SFT plays a crucial role in transforming a model's general knowledge into structured responses tailored to specific tasks. However, there is no clearly established methodology for effective training data selection. Simply increasing the volume of data does not guarantee performance improvements, while preprocessing, sampling, and validation require substantial time and cost.
To address this issue, a variety of data selection methods have been proposed. Among them, knowledge based selection approaches identify suitable training data by analyzing the model's responses. Nevertheless, these methods typically rely on prompt engineering, making them sensitive to variations and incurring additional costs for prompt design.
In this study, we propose Knowledge Analysis via Model Internal Representations (KAMIR), a novel approach that overcomes these limitations by analyzing data based on the model's internal representations. KAMIR computes similarities between the hidden states of each layer (block) and the final hidden states for a given input to assess the data. Unlike prior methods that were largely limited to multiple choice tasks, KAMIR can be applied to a wide range of tasks such as machine reading comprehension and summarization. Moreover, it selects data useful for training based on the model's familiarity with the input, even with a small dataset and a simple classifier architecture. Experiments across diverse task datasets demonstrate that training with less familiar data leads to better generalization performance.

\end{abstract}

\section{Introduction}

Various training methodologies have been developed to enhance the performance of large language models (LLM) and to enable them to perform a wide range of tasks. Broadly, these training processes can be categorized into pretraining, supervised fine tuning (SFT), and alignment tuning\cite{llm_train}. Among these, SFT is the process of refining the general knowledge acquired during pretraining so that the model can produce structured outputs tailored to specific tasks. To achieve this, high quality training data that accurately reflect the characteristics of the target task are required.

However, there is still no definitive solution to the problem of selecting effective training data for SFT. In most cases, researchers and developers must rely on trial and error to identify the optimal data composition. When handling large scale datasets often numbering in the millions the preprocessing, sampling, and validation processes demand considerable time and effort. Furthermore, simply increasing the volume of data does not guarantee performance improvements; on the contrary, the inclusion of redundant or low quality samples may reduce training efficiency. Empirical studies suggest that datasets containing tens of millions of examples are often required to achieve significant performance gains, which entails substantial costs and time.

For these reasons, recent research has actively explored efficient data selection methods, such as importance based sampling, representative sample selection via clustering, and uncertainty based augmentation. These approaches have shown potential in achieving strong performance with smaller amounts of data.

Among them, knowledge based detection methods select training data not by intrinsic data properties but by the model's responses to the data. Prior research has demonstrated that such methods can improve model training, for example, by showing that data consistent with knowledge acquired during pretraining even if incorrect can still benefit the model, or that training on data robust to prompt bias can enhance performance on domain specific knowledge. Nonetheless, these approaches heavily depend on prompt engineering, making them sensitive to minor variations and largely limited to tasks such as question answering (QA), where ground truth verification is straightforward.

To address these limitations, we propose Knowledge Analysis via Model Internal Representations (KAMIR), a method that analyzes data through the model's internal representations without requiring prompt related manipulations such as task descriptions or additional exemplars.

KAMIR leverages the concept of the logit lens by utilizing hidden states produced at each layer or block, as well as the final hidden state that encapsulates the model's full interpretation of the input\cite{logitlens}. This enables us to track how the model processes and interprets data across layers and to analyze data based on these representational dynamics.

In practice, we collected both widely known knowledge and post deployment data, analyzed them through KAMIR, and employed a simple classifier trained on these analyses for data categorization and learning. The results demonstrate that models trained on data categorized alongside temporally inaccessible knowledge such as newly acquired data achieve superior performance compared to those trained without such categorization. This finding validates that analyzing data via internal representations enables the selection of training data that improve generalization performance.

The contributions of this study are as follows:
\begin{enumerate}

\item We propose a robust data analysis method based on internal representations of the model, free from prompt dependency.

\item We extend data analysis beyond QA to a variety of tasks.

\item We demonstrate that even with a small amount of data and a simple classifier, it is possible to effectively select training data that improve model performance.
\end{enumerate}

\section{Related Works}
Self-Align explores the parameter knowledge of a pretrained LLM through few shot in context learning (ICL) and subsequently constructs Instruction Fine Tuning (IFT) datasets aligned with this internal knowledge\cite{self-align}. By doing so, it maintains consistency between the model's internal knowledge and the provided instructions, facilitating knowledge detection and alignment.

KaFT (Knowledge aware Fine Tuning) is a knowledge aware fine tuning method designed to enhance an LLM's domain specific question answering performance\cite{kaft}. KaFT employs ICL while accounting for positional bias and assigns rewards differently based on the level of knowledge conflict in training samples, enabling the classification of known data.

KGLens evaluates the alignment between an LLM and a knowledge graph (KG) by generating natural language questions from the KG and using a structure based importance sampling strategy to efficiently detect the model's knowledge\cite{kglens}.

Jiang et al. highlighted the limitations of simple cloze pattern prompts for extracting knowledge from LLM\cite{knowledge_probing_1}. To address this, they proposed mining based and paraphrasing based automatic prompt generation methods, which more precisely elicit knowledge and cover diverse expressions.

Tighidet et al. analyzed whether an LLM relies on parametric knowledge (PK) or contextual knowledge (CK) by employing prompts containing information conflicting with the model's PK and examining the resulting internal activations\cite{knowledge_probing_2}. This approach helps to understand the model's reliance on internal versus contextual knowledge sources.

\section{KAMIR : Knowledge Analysis via Model Internal Representations}

\begin{figure*}[thb!]
\centering
\includegraphics[width=\textwidth]{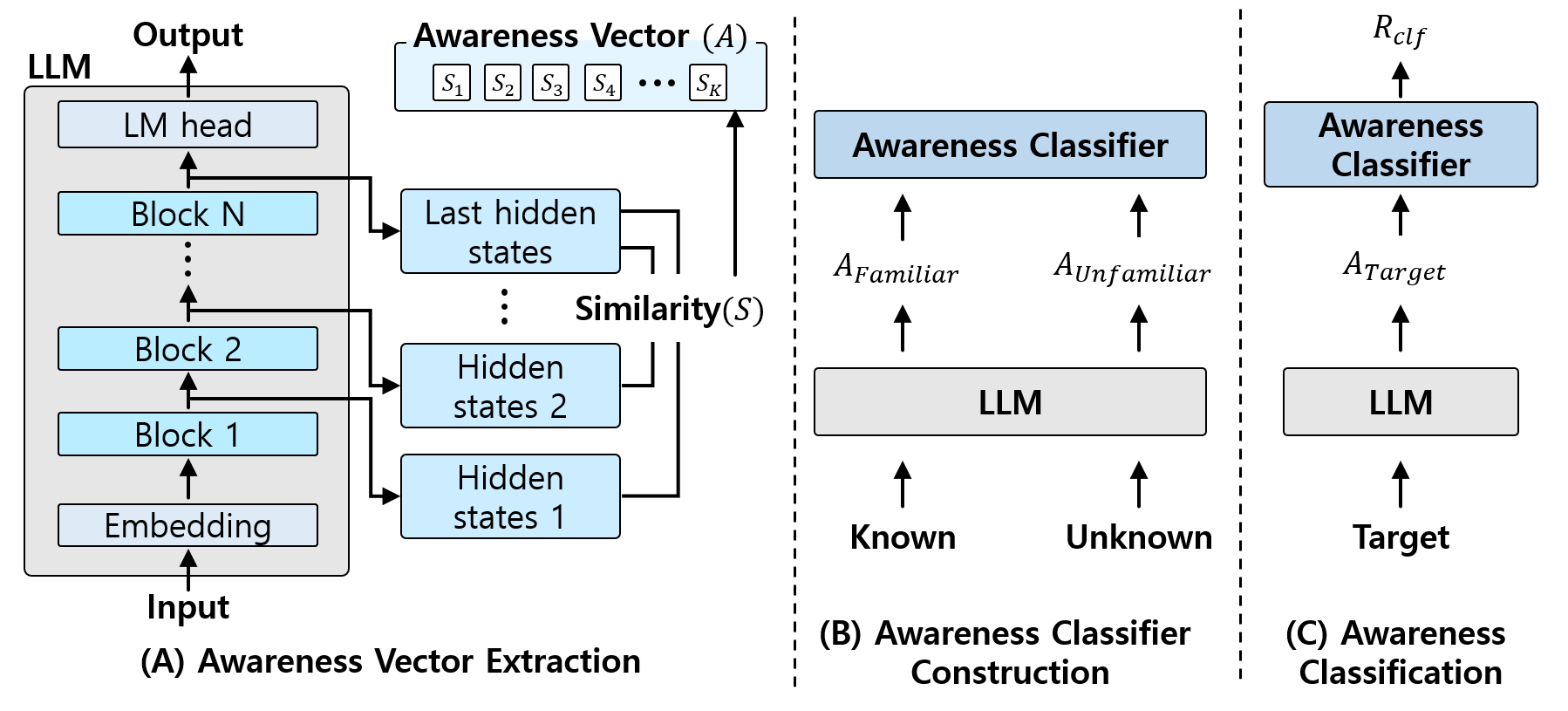}
\caption{Extracting awareness vector(A), constructing awareness classifier (B) and classifying model's awareness of target data (C) with Knowledge Analysis via Model Internal Representations(KAMIR)}
\label{fig:fig1}
\end{figure*} 

\subsection{Awareness Vector Extraction}
As a model processes input data, it passes through multiple layers (blocks), each analyzing the data in different ways. In this study, we measure the model's awareness of the input data by examining how the representation vectors evolve across layers specifically, whether the analysis remains consistent or diverges at different layers. The procedure for computing the model's awareness of a given input is illustrated in Figure 1(A).
First, the input is provided to the model, which generates the corresponding output. At this stage, only the raw content is used, without additional task related descriptions or answer options. Next, we collect the hidden states($H$) from each layer at the time of generating the final token. We restrict collection to the final token because its representation vector encapsulates information from the input, the tokens generated prior to the final token, and the final token itself.
Finally, we compute the similarity($S$) between the hidden states of each intermediate layer and the last hidden state of the final layer. Cosine similarity is employed for this purpose. The collection of these similarity scores is then defined as the model's awareness vector($A$) for the given input.

$$A = \begin{bmatrix} S_1 & S_2 & \cdots & S_K \end{bmatrix}^{\!\top},
\quad 
S_a = \frac{H_a \cdot H_K}{\lVert H_a \rVert \,\lVert H_K \rVert}, 
\quad K = N-1$$

\begin{figure}[thb!]
\centering
\includegraphics[width=\textwidth]{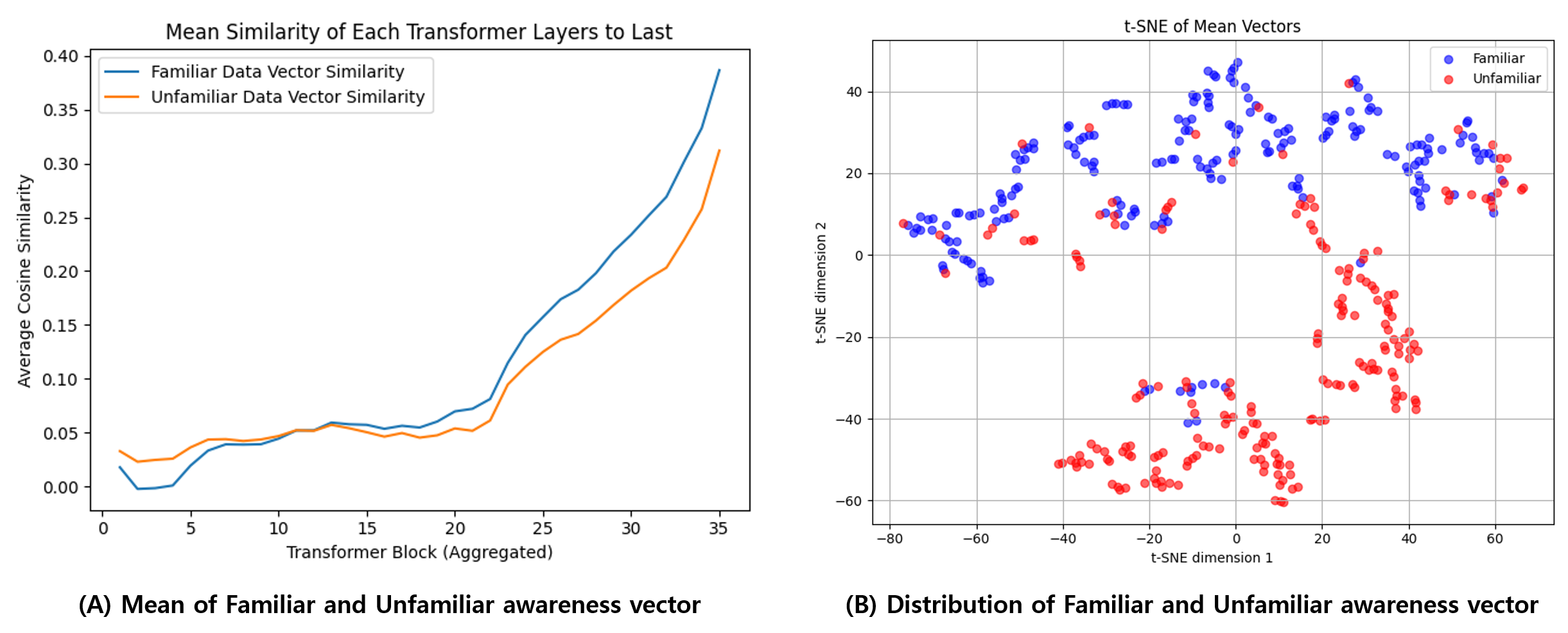}
\caption{Mean(A) and Distribution(B) of Awarness Vector used for constructing Data Awareness Classifier}
\label{fig:fig2}
\end{figure}

\subsection{Data Awareness Classifier}
Based on the awareness vector, we constructed an MLP based awareness classifier, as illustrated in Figure 1(B).
Ideally, to accurately distinguish between data that an LLM has learned and data it has not, one would control the pretraining stage itself. However, since this study focuses on evaluating performance in commonly used open models, we instead collected data based on general awareness: data that the model was highly likely to have learned (familiar data) and data it was unlikely to have learned (unfamiliar data).
For familiar data, we collected document samples concerning well known events, figures, and companies that occurred prior to the model's release date. For unfamiliar data, we gathered documents on distinctive events, newly released films, and scientific papers published after the model's release. While it is straightforward to assume that post release events or publications were not included in pretraining, such data may still partially overlap with prior knowledge through shared entities, similar event patterns, basic reasoning, or background knowledge. Thus, completely unlearned data are difficult to identify. To address this, we focused on collecting data that would be less inferable from prior knowledge and thus less familiar to the model.
Each collected dataset was divided into sub passages of a fixed token length, and awareness vectors were computed for each sub passage using the method described in Section 3.1. The average of these sub passage awareness vectors was then taken as the awareness vector for the entire dataset.
Finally, using the awareness vectors of familiar and unfamiliar datasets ($A_{Familiar}$, $A_{Unfamiliar}$), we trained a data awareness classifier composed of a simple MLP layer. As illustrated in Figure 1(C), this classifier takes as input the awareness vector of a new dataset ($A_{Target}$) and classifies it as either familiar or unfamiliar ($R_{clf}$).
Figure 2(A) shows the average awareness vectors of familiar and unfamiliar datasets. Both vectors exhibit a similar trend of increases and decreases across layers, suggesting that the model processes data in a comparable manner regardless of familiarity. However, the magnitude of activation differs significantly depending on whether the data are familiar or unfamiliar, highlighting awareness dependent variation.
Figure 2(B) presents the distribution of familiar and unfamiliar datasets visualized via t-SNE. Although some overlap exists, the clusters are sufficiently distinct to allow meaningful classification even by visual inspection. This result indicates that awareness vectors share consistent characteristics within each class, thereby enabling reliable classification of new data.

\section{Experiments}
In this section, we analyze the experimental results regarding the impact of intrinsic knowledge detection on training performance.

\subsection{Experimental Setup}

To measure SFT performance with respect to familiar and unfamiliar data, we used a pretrained only model. Specifically, we adopted Qwen3-4B-Base as the base model and employed LoRA for fine tuning\cite{qwen3}.
In addition to familiar and unfamiliar datasets, we included a randomly sampled dataset of equal size to establish a control condition comparable to common data selection practices.
For awareness vector computation, the maximum input length was set to 300 tokens, and the output length was limited to 100 tokens.
We employed training and evaluation datasets spanning diverse domains and tasks. 
SQuAD 1.1 is an english reading comprehension dataset where answers are extractable directly from documents\cite{squad1.1}. TriviaQA is a large scale QA dataset consisting of questions and answers collected from the web across multiple domains\cite{triviaqa}. KorQuAD 1.1 is a Korean reading comprehension dataset constructed in the SQuAD format. MedQA is a QA dataset based on specialized medical knowledge and KorMedMCQA is a Korean multiple choice dataset in the medical domain\cite{medqa,kormedmcqa}. SciQ is a dataset containing science related questions and answers designed for primary and secondary education\cite{sciq}. For summarization task, we use XLSum and CNN/DailyMail, which is multilingual document summarization and news summarization datasets primarly constructed for MRC respectively\cite{xlsum,cnndailymail}.
For evaluation, we considered both the training datasets and additional benchmark datasets. For machine reading comprehension, we used SQuAD 1.1, SQuAD 2.0, and TriviaQA, as well as MedQA, MedMCQA, and SciQ\cite{squad2.0}. For summarization evaluation, we used XLSum and CNN/DailyMail. This setup enabled us to comprehensively assess model performance across diverse domains and tasks.

The evaluation metrics were as follows:
\begin{itemize}
    \item SQuAD 1.1 and SQuAD 2.0: F1 score
    \item TriviaQA: exact match (ignoring whitespace)
    \item MedQA, MedMCQA, SciQ: accuracy
    \item XLSum, CNN/DailyMail: pairwise comparison using GPT-4o-mini, comparing outputs trained on unfamiliar data against those trained on familiar or randomly sampled data.
\end{itemize}

This comprehensive evaluation framework allowed us to assess the model's performance across a wide range of tasks and domains.

\subsection{Training Effects of Familiar vs. Unfamiliar Data}
\begin{table}[]
\centering
\caption{Evaluation result for each dataset}
\label{Table1}
\resizebox{\textwidth}{!}{%
\begin{tabular}{ccc|cccc}
\hline
Train data                 & Number of data         & Test data   & Base    & Familiar                                                                    & Unfamiliar & Random                                                                      \\ \hline
                           &                        &             & \multicolumn{4}{c}{MRC}                                                                                                                                                          \\ \hline
\multirow{2}{*}{SQuAD 1.1} & \multirow{2}{*}{34443} & SQuAD  1.1  & 72.6789 & 62.8581                                                                     & 78.4130    & 71.2798                                                                     \\
                           &                        & Squad 2.0   & 35.9541 & 31.3089                                                                     & 39.1838    & 35.5285                                                                     \\
TriviaQA                   & 31800                  & TriviaQA    & 0.4063  & 0.4423                                                                      & 0.4800     & 0.4407                                                                      \\
KorQuAD 1.1                & 21021                  & KorQuAD 1.1 & 68.9435 & 85.5205                                                                     & 88.4342    & 88.1442                                                                     \\ \hline
                           &                        &             & \multicolumn{4}{c}{MCQA}                                                                                                                                                         \\ \hline
\multirow{2}{*}{MedQA}     & \multirow{2}{*}{850}   & MedQA       & 0.6190  & 0.6159                                                                      & 0.6190     & 0.6143                                                                      \\
                           &                        & MedMCQA    & 0.5721  & 0.5747                                                                      & 0.5680     & 0.5723                                                                      \\
SciQ                       & 3100                   & SciQ        & 0.966   & 0.914                                                                       & 0.926      & 0.919                                                                       \\
KorMedMCQA                 & 1070                   & KorMedMCQA  & 0.0705  & 0.4955                                                                      & 0.5101     & 0.5188                                                                      \\ \hline
                           &                        &             & \multicolumn{4}{c}{SMR}                                                                                                                                                          \\ \hline
XLSum                      & 19000                  & -           & -       & \begin{tabular}[c]{@{}c@{}}9.2\% | 7.1\% | 83.7\%\end{tabular}  & -          & \begin{tabular}[c]{@{}c@{}}11.7\% | 6.6\% | 81.7\%\end{tabular} \\
CNN/Dailymail              & 19100                  & -           & -       & \begin{tabular}[c]{@{}c@{}}50.3\% | 7.9\% | 41.9\%\end{tabular} & -          & \begin{tabular}[c]{@{}c@{}}51.4\% | 8.0\% | 40.6\%\end{tabular} \\
XLSum\_ko                  & 1614                   & -           & -       & \begin{tabular}[c]{@{}c@{}}7.8\% | 10.0\% | 82.2\%\end{tabular} & -          & \begin{tabular}[c]{@{}c@{}}7.5\% | 7.1\% | 85.4\%\end{tabular}  \\ \hline
\end{tabular}%
}
\end{table}

The comparative performance of models trained with familiar, unfamiliar, and randomly sampled data is summarized in Table 1. Across most datasets, models trained with unfamiliar data consistently outperformed those trained with familiar data.

In the machine reading comprehension (MRC) domain, the unfamiliar trained model outperformed the familiar trained model across all datasets. Notably, on the SQuAD series, the familiar trained model suffered a marked decline in performance, whereas the unfamiliar trained model achieved improvements. A similar performance gain was also observed on TriviaQA. These results suggest that unfamiliar data provide richer contexts and more diverse question types, thereby enhancing the model's ability for answer localization and contextual comprehension.
In the multiple choice QA (MCQA) domain, despite the limited training data size, the performance drop of the unfamiliar trained model was comparatively smaller than that of the familiar trained model. In most evaluations, the unfamiliar trained model achieved superior performance. This indicates that distributional diversity within unfamiliar data strengthened the model's generalization ability, particularly in specialized domains such as medicine and science.
In the summarization (SMR) domain, results varied by dataset. On XLSum and XLSum\_ko, the quality difference between familiar trained and unfamiliar trained models was marginal, with more than 80\% of outcomes resulting in ties. This reflects the inherently high variability of valid outputs in summarization tasks, where multiple expressions can serve as correct answers, thereby limiting the effect of unfamiliar training on evaluation quality.
In contrast, on CNN/DailyMail, although approximately 40\% of outcomes were ties, both familiar and random trained models won in over 50\% of cases. This can be attributed to the dataset's extractive summarization nature: unfamiliar training may increase output diversity, which in turn leads to discrepancies with the reference distribution, ultimately reducing performance.
These observations held not only for English datasets but also for Korean datasets such as XLSum\_ko, indicating the generalizability of the findings across languages.

\subsection{Analysis of Training Effects}

\begin{table}[]
\centering
\caption{Loss, Entory, Gradient norm per dataset}
\label{Table2}
\resizebox{\textwidth}{!}{%
\begin{tabular}{ccc|ccc}
\hline
\multicolumn{1}{c}{Task Type} & \multicolumn{1}{c}{Dataset} & Group & Loss & Entropy & \multicolumn{1}{c}{Grad norm} \\ \hline
\multirow{6}{*}{MRC} & \multirow{2}{*}{SQuAD 1.1}          & Known   & 13.5519 & 2.0498 & 11.9417 \\
                     &                                 & Unknown & 13.4033 & 2.1139 & 12.5196 \\ \cline{2-6} 
                     & \multirow{2}{*}{TriviaQA}       & Known   & 13.6715 & 1.7530 & 7.9932  \\
                     &                                 & Unknown & 13.4039 & 1.9149 & 8.5120  \\ \cline{2-6} 
                     & \multirow{2}{*}{KorQuAD 1.1}        & Known   & 12.5260 & 2.0840 & 7.8656  \\
                     &                                 & Unknown & 12.4629 & 2.1049 & 7.8961  \\ \hline
\multirow{6}{*}{QA}  & \multirow{2}{*}{MedQA}          & Known   & 13.7439 & 1.6533 & 9.1848  \\
                     &                                 & Unknown & 13.8006 & 1.6878 & 9.4314  \\ \cline{2-6} 
                     & \multirow{2}{*}{SciQ}           & Known   & 13.9105 & 1.7186 & 13.9221 \\
                     &                                 & Unknown & 13.7475 & 1.7327 & 13.8577 \\ \cline{2-6} 
                     & \multirow{2}{*}{KorMedMCQA}     & Known   & 13.1609 & 1.8022 & 11.1142 \\
                     &                                 & Unknown & 12.9098 & 1.8857 & 10.9080 \\ \hline
\multirow{6}{*}{SMR} & \multirow{2}{*}{XL-Sum}          & Known   & 13.1334 & 2.3662 & 10.1204 \\
                     &                                 & Unknown & 13.1104 & 2.4081 & 10.6592 \\ \cline{2-6} 
                     & \multirow{2}{*}{CNN/Dailymail} & Known   & 12.9277 & 2.2566 & 9.3773  \\
                     &                                 & Unknown & 13.4547 & 2.2919 & 9.4190  \\ \cline{2-6} 
                     & \multirow{2}{*}{XL-Sum\_ko}      & Known   & 13.0744 & 1.9869 & 7.8703  \\
                     &                                 & Unknown & 13.0428 & 2.0186 & 8.3573  \\ \hline
\end{tabular}%
}
\end{table}

In this study, we analyzed performance differences according to the composition of training data (Familiar, Unfamiliar, and Random) and examined the underlying causes from the perspectives of loss, entropy, and gradient norm.

Overall, unfamiliar data maintained loss values that were generally lower or comparable to those of familiar data, indicating stable convergence during training. Moreover, the prediction distributions for unfamiliar data exhibited relatively higher entropy, suggesting that the model formed more generalized probability distributions rather than being overly confident in specific answers. This increase in uncertainty can help mitigate overfitting and enhance adaptability to diverse input distributions. Additionally, gradient norms were generally higher when training with unfamiliar data, implying more active exploration of the parameter space, which likely contributed to improved generalization performance.

In MCQA and MRC tasks, where answers are concise and clearly defined, training with unfamiliar data enhanced the model's ability for answer inference and contextual comprehension. By exposing the model to diverse question types and contextual structures, unfamiliar training reduced overfitting and promoted distributional generalization.

Conversely, in SMR tasks, although models trained on unfamiliar data demonstrated favorable loss and entropy metrics on XLSum and XLSum\_ko, the LLM-as-a-judge evaluation revealed minimal quality differences between familiar and unfamiliar training, with the majority of comparisons resulting in ties. This outcome reflects the inherent multi reference nature of summarization tasks and their relative evaluation scheme, where multiple valid outputs are possible, thereby limiting the observable effect of unfamiliar training on evaluation.

In CNN/DailyMail, unfamiliar training led to substantially higher loss and inferior evaluation results compared to familiar and random trained models. This dataset is inherently extractive, where reference sentences explicitly exist within the source text. In such settings, strategies emphasizing diversity and distributional generalization characteristic of unfamiliar training can be disadvantageous. While unfamiliar training encouraged the model to generate varied candidate sentences, this generative diversity increased answer distribution discrepancies, resulting in higher loss and decreased evaluation performance.

Although MRC is also extractive, answers are confined to short spans. Consequently, the generalization benefits of unfamiliar data were positively realized: higher entropy and active parameter exploration allowed the model to reliably identify correct spans even in novel contexts.

In summary, unfamiliar data training contributes to improved generalization performance through:
\begin{itemize}
    \item Stabilized convergence (reduced loss)
    \item Increased prediction uncertainty (higher entropy)
    \item Enhanced parameter space exploration (increased gradient norm)
\end{itemize}

However, the effectiveness depends on answer length, answer variability, data structure, and evaluation characteristics. Its impact was most pronounced in tasks with concise, unambiguous answers (MRC, MCQA), limited in generation based summarization tasks (XLSum/XLSum\_ko), and even detrimental for long, extractive summaries (CNN/DailyMail) due to answer distribution discrepancies. Thus, the utility of unfamiliar data is not solely determined by answer clarity but by the interaction between answer characteristics, data structure, and evaluation methodology.

\section{Conclusions}
In this study, we proposed Knowledge Analysis via Model Internal Representations(KAMIR), a method for detecting intrinsic knowledge in LLMs without relying on prompts. KAMIR computes awareness vectors by measuring the similarity between hidden states at each model layer and the final output vector, which are then used to construct a data awareness classifier distinguishing Familiar and Unfamiliar data. This approach overcomes the limitations of prompt based intrinsic knowledge detection, including sensitivity and multiple choice task constraints.
Experimental results demonstrated that the proposed method effectively differentiates familiar and unfamiliar data across diverse tasks beyond multiple choice, including MRC and summarization. Notably, SFT training with unfamiliar data achieved higher performance than familiar data across most datasets. This improvement was linked to reduced loss, increased prediction entropy, and greater gradient norms during training, indicating enhanced generalization. The effects were particularly pronounced in tasks with concise, well defined answers, such as MRC and MCQA, whereas for generation based tasks with inherently variable outputs, such as summarization, performance improvements were comparatively limited.
Future work includes extending the approach to additional languages and domains, integrating awareness based sampling and clustering to develop more efficient SFT strategies, and exploring evaluation and training methods to maximize the benefits of unfamiliar data in generative tasks. This study offers a novel perspective on LLM training data selection and intrinsic knowledge utilization, demonstrating the potential for efficient and generalizable model training.

\newpage
\bibliographystyle{unsrt}
\bibliography{ref}

\begin{thebibliography}{10}

\bibitem{llm_train}
Hanyu Lai, Xiao Liu, Junjie Gao, Jiale Cheng, Zehan Qi, Yifan Xu, Shuntian Yao, Dan Zhang, Jinhua Du, Zhenyu Hou, Xin Lv, Minlie Huang, Yuxiao Dong, and Jie Tang.
\newblock A survey of post-training scaling in large language models.
\newblock In Wanxiang Che, Joyce Nabende, Ekaterina Shutova, and Mohammad~Taher Pilehvar, editors, {\em Proceedings of the 63rd Annual Meeting of the Association for Computational Linguistics (Volume 1: Long Papers)}, pages 2771--2791, Vienna, Austria, July 2025. Association for Computational Linguistics.

\bibitem{logitlens}
nostalgebraist.
\newblock Interpreting gpt: The logit lens.
\newblock \url{https://www.lesswrong.com/posts/AcKRB8wDpdaN6v6ru/interpreting-gpt-the-logit-lens}, 2020.

\bibitem{self-align}
Mengjie Ren, Boxi Cao, Hongyu Lin, Cao Liu, Xianpei Han, Ke~Zeng, Wan Guanglu, Xunliang Cai, and Le~Sun.
\newblock Learning or self-aligning? rethinking instruction fine-tuning.
\newblock In Lun-Wei Ku, Andre Martins, and Vivek Srikumar, editors, {\em Proceedings of the 62nd Annual Meeting of the Association for Computational Linguistics (Volume 1: Long Papers)}, pages 6090--6105, Bangkok, Thailand, August 2024. Association for Computational Linguistics.

\bibitem{kaft}
Qihuang Zhong, Liang Ding, Xiantao Cai, Juhua Liu, Bo~Du, and Dacheng Tao.
\newblock {K}a{FT}: Knowledge-aware fine-tuning for boosting {LLM}s' domain-specific question-answering performance.
\newblock In Wanxiang Che, Joyce Nabende, Ekaterina Shutova, and Mohammad~Taher Pilehvar, editors, {\em Findings of the Association for Computational Linguistics: ACL 2025}, pages 24085--24100, Vienna, Austria, July 2025. Association for Computational Linguistics.

\bibitem{kglens}
Shangshang Zheng, He~Bai, Yizhe Zhang, Yi~Su, Xiaochuan Niu, and Navdeep Jaitly.
\newblock Kglens: Towards efficient and effective knowledge probing of large language models with knowledge graphs, 2024.

\bibitem{knowledge_probing_1}
Zhengbao Jiang, Frank~F. Xu, Jun Araki, and Graham Neubig.
\newblock How can we know what language models know?
\newblock {\em Transactions of the Association for Computational Linguistics}, 8:423--438, 2020.

\bibitem{knowledge_probing_2}
Zineddine Tighidet, Jiali Mei, Benjamin Piwowarski, and Patrick Gallinari.
\newblock Probing language models on their knowledge source.
\newblock In Yonatan Belinkov, Najoung Kim, Jaap Jumelet, Hosein Mohebbi, Aaron Mueller, and Hanjie Chen, editors, {\em Proceedings of the 7th BlackboxNLP Workshop: Analyzing and Interpreting Neural Networks for NLP}, pages 604--614, Miami, Florida, US, November 2024. Association for Computational Linguistics.

\bibitem{qwen3}
An~Yang, Anfeng Li, Baosong Yang, Beichen Zhang, Binyuan Hui, Bo~Zheng, Bowen Yu, Chang Gao, Chengen Huang, Chenxu Lv, Chujie Zheng, Dayiheng Liu, Fan Zhou, Fei Huang, Feng Hu, Hao Ge, Haoran Wei, Huan Lin, Jialong Tang, Jian Yang, Jianhong Tu, Jianwei Zhang, Jianxin Yang, Jiaxi Yang, Jing Zhou, Jingren Zhou, Junyang Lin, Kai Dang, Keqin Bao, Kexin Yang, Le~Yu, Lianghao Deng, Mei Li, Mingfeng Xue, Mingze Li, Pei Zhang, Peng Wang, Qin Zhu, Rui Men, Ruize Gao, Shixuan Liu, Shuang Luo, Tianhao Li, Tianyi Tang, Wenbiao Yin, Xingzhang Ren, Xinyu Wang, Xinyu Zhang, Xuancheng Ren, Yang Fan, Yang Su, Yichang Zhang, Yinger Zhang, Yu~Wan, Yuqiong Liu, Zekun Wang, Zeyu Cui, Zhenru Zhang, Zhipeng Zhou, and Zihan Qiu.
\newblock Qwen3 technical report, 2025.

\bibitem{squad1.1}
Pranav Rajpurkar, Jian Zhang, Konstantin Lopyrev, and Percy Liang.
\newblock {SQ}u{AD}: 100,000+ questions for machine comprehension of text.
\newblock In Jian Su, Kevin Duh, and Xavier Carreras, editors, {\em Proceedings of the 2016 Conference on Empirical Methods in Natural Language Processing}, pages 2383--2392, Austin, Texas, November 2016. Association for Computational Linguistics.

\bibitem{triviaqa}
Mandar Joshi, Eunsol Choi, Daniel Weld, and Luke Zettlemoyer.
\newblock {T}rivia{QA}: A large scale distantly supervised challenge dataset for reading comprehension.
\newblock In Regina Barzilay and Min-Yen Kan, editors, {\em Proceedings of the 55th Annual Meeting of the Association for Computational Linguistics (Volume 1: Long Papers)}, pages 1601--1611, Vancouver, Canada, July 2017. Association for Computational Linguistics.

\bibitem{medqa}
Di~Jin, Eileen Pan, Nassim Oufattole, Wei-Hung Weng, Hanyi Fang, and Peter Szolovits.
\newblock What disease does this patient have? a large-scale open domain question answering dataset from medical exams.
\newblock {\em Applied Sciences}, 11(14), 2021.

\bibitem{kormedmcqa}
Sunjun Kweon, Byungjin Choi, Gyouk Chu, Junyeong Song, Daeun Hyeon, Sujin Gan, Jueon Kim, Minkyu Kim, Rae~Woong Park, and Edward Choi.
\newblock Kormedmcqa: Multi-choice question answering benchmark for korean healthcare professional licensing examinations, 2024.

\bibitem{sciq}
Johannes Welbl, Nelson~F. Liu, and Matt Gardner.
\newblock Crowdsourcing multiple choice science questions.
\newblock In Leon Derczynski, Wei Xu, Alan Ritter, and Tim Baldwin, editors, {\em Proceedings of the 3rd Workshop on Noisy User-generated Text}, pages 94--106, Copenhagen, Denmark, September 2017. Association for Computational Linguistics.

\bibitem{xlsum}
Tahmid Hasan, Abhik Bhattacharjee, Md.~Saiful Islam, Kazi Mubasshir, Yuan-Fang Li, Yong-Bin Kang, M.~Sohel Rahman, and Rifat Shahriyar.
\newblock {XL}-sum: Large-scale multilingual abstractive summarization for 44 languages.
\newblock In Chengqing Zong, Fei Xia, Wenjie Li, and Roberto Navigli, editors, {\em Findings of the Association for Computational Linguistics: ACL-IJCNLP 2021}, pages 4693--4703, Online, August 2021. Association for Computational Linguistics.

\bibitem{cnndailymail}
Danqi Chen, Jason Bolton, and Christopher~D. Manning.
\newblock A thorough examination of the {CNN}/{D}aily {M}ail reading comprehension task.
\newblock In Katrin Erk and Noah~A. Smith, editors, {\em Proceedings of the 54th Annual Meeting of the Association for Computational Linguistics (Volume 1: Long Papers)}, pages 2358--2367, Berlin, Germany, August 2016. Association for Computational Linguistics.

\bibitem{squad2.0}
Pranav Rajpurkar, Robin Jia, and Percy Liang.
\newblock Know what you don{'}t know: Unanswerable questions for {SQ}u{AD}.
\newblock In Iryna Gurevych and Yusuke Miyao, editors, {\em Proceedings of the 56th Annual Meeting of the Association for Computational Linguistics (Volume 2: Short Papers)}, pages 784--789, Melbourne, Australia, July 2018. Association for Computational Linguistics.

\end{thebibliography}

\end{document}